\documentclass[a4paper,11pt,article,oneside]{memoir}
\usepackage{ecis2024}

\maintitle{IGANN Sparse: Bridging Sparsity and Interpretability with Non-linear Insight} 
\shorttitle{IGANN Sparse} 
\category{Research in Progress} 

\authors{
Stoecker, Theodor Felix, FAU Erlangen-Nürnberg, Nürnberg, Germany, theo.stoecker@fau.de

Hambauer, Nico, FAU Erlangen-Nürnberg, Nürnberg, Germany, nico.hambauer@fau.de

Zschech, Patrick, Leipzig University, Leipzig, Germany, patrick.zschech@uni-leipzig.de

Kraus, Mathias, University of Regensburg, Regensburg, Germany, mathias.kraus@ur.de}

\shortauthors{Stoecker et al.} 

\usepackage{array}
\usepackage{hyperref}
\usepackage{url}
\usepackage{graphicx}
\usepackage{algorithm}
\usepackage{algorithmic}
\usepackage{todonotes}
\usepackage{subcaption} 

\usepackage{siunitx}

\begin{document}

\begin{abstract}\noindent
Feature selection is a critical component in predictive analytics that significantly affects the prediction accuracy and interpretability of models. 
Intrinsic methods for feature selection are built directly into model learning, providing a fast and attractive option for large amounts of data.
Machine learning algorithms, such as penalized regression models (e.g., lasso) are the most common choice when it comes to in-built feature selection.
However, they fail to capture non-linear relationships, which ultimately affects their ability to predict outcomes in intricate datasets.
In this paper, we propose IGANN Sparse, a novel machine learning model from the family of generalized additive models, which promotes sparsity through a non-linear feature selection process during training.
This ensures interpretability through improved model sparsity without sacrificing predictive performance. 
Moreover, IGANN Sparse serves as an exploratory tool for information systems researchers to unveil important non-linear relationships in domains that are characterized by complex patterns.
Our ongoing research is directed at a thorough evaluation of the IGANN Sparse model, including user studies that allow to assess how well users of the model can benefit from the reduced number of features. 
This will allow for a deeper understanding of the interactions between linear vs. non-linear modeling, number of selected features, and predictive performance.
\end{abstract}


\chapter{Introduction}

Predictive analytics is a crucial methodological stream of research in the field of information systems (IS) that deals with the creation of empirical prediction models \citep{shmueli_predictive_2011}. By leveraging advanced machine learning (ML) techniques, researchers can uncover patterns and relationships within large datasets, enabling them to anticipate future events, user behaviors, and system performance \citep{kuhl_how_2021}. In addition to its practical utility, predictive analytics also plays an important role in theory development and testing, as well as relevance assessment \citep{shmueli_predictive_2011}.

In recent years, the IS community has increasingly recognized the importance of ensuring that prediction models not only provide high predictive performance, but are also comprehensible for explanation purposes \citep{bauer_explained_2023, kim_rolex_2023}. There are basically two primary approaches for ensuring model explainability: inherently interpretable models and post-hoc explainability methods.\footnote{It should be noted that the terms \emph{interpretability} and \emph{explainability} can have different meanings from a psychological perspective \citep[cf.][]{broniatowski}. In this paper, however, we will concentrate on a technical differentiation \citep[cf.][]{IGANN_Kraus}.} Inherently interpretable models, such as shallow decision trees, linear models, and generalized additive models (GAMs) \citep{GAM_Hastie_Tibishirani}, are designed to be human-readable without requiring additional explanation. These models often employ techniques like linearity, and sparsity to enhance their interpretability \citep{rudin_stop_2019}. On the other hand, post-hoc explainability tools like Shapley values \citep{lundberg2017unified} or LIME \citep{ribeiro2016should} aim to approximate the complex behavior of non-interpretable models \citep{esteva_guide_2019}. However, these post-hoc methods should be applied cautiously, as they may not fully capture the intricate workings of the original models. For tabular data, the choice between the two approaches is often guided by the marginal performance gains achievable with complex models over interpretable ones \citep{rudin_stop_2019, zschech2022gam}.

GAMs and their various extensions base their final prediction on a number of independent functions that map the input features to the output space of the model, where they are summed up to generate the final prediction \citep{GAM_Hastie_Tibishirani}. Each of these functions usually only processes an individual feature or a single interaction between two features, which allows to visualize the function after training and gaining insights into the effect that a feature has on the model output.

Although GAMs have proven to be very powerful, they lack an important property that limits their applicability to high-dimensional feature spaces (i.e., datasets characterized by a large number of input features): They do not easily allow the creation of sparse models, i.e., models that base their predictions on only a few input features. This is the case because GAMs typically are trained in an iterative fashion (through backfitting or gradient boosting), where a feature is selected in each iteration to minimize the remaining loss of the model \citep{GAM_Hastie_Tibishirani, GAM_EBM}. Although very powerful, this iterative approach makes the implementation of additional sparsity constraints difficult. Consequently, commonly used are linear feature selection methods, however data can have non-linear relations, which simple models such as a linear regressions are unable to detect.

Only recently, novel neural network based GAMs have been proposed which do not rely on the iteration over features \citep[e.g.,][]{GAM_GAMI-Net, IGANN_Kraus}. This paper extends the Interpretable Generalized Additive Neural Network (IGANN) framework with a focus on sparsity and interpretability. It also positions the IGANN sparse model as a powerful exploratory tool for exploring non-linear dependencies in data that traditional GAMs may not readily uncover.

Our contributions are fourfold: First, we propose a novel approach to fast training of sparse neural networks using extreme learning machines. Second, we incorporate this method into the IGANN model, resulting in the introduction of IGANN Sparse.\footnote{mplementation on GitHub: https://github.com/MathiasKraus/igann}  Third, we validate that IGANN Sparse maintains comparable predictive accuracy to its non-sparse counterparts while significantly reducing the number of features, thereby improving interpretability. Finally, we demonstrate the utility of IGANN Sparse in non-linear feature selection, establishing its role in exploratory data analysis and interpretative modeling.

Our work has implications for both predictive analytics research and practice. We address the issue of feature selection, which is a common step during pre-processing of machine learning pipelines. To achieve this, we expand IGANN, a model capable of learning non-linear relations in data, by presenting a sparse IGANN version. Furthermore, our approach has implications for IS researchers, as training a sparse model for feature selection is a promising way to keep the logic of prediction models comprehensible. Therefore, our model offers a promising tool for empirical IS studies that are concerned with popular research questions related to predictive model building, such as predicting purchase behavior, price dynamics, user satisfaction or technology acceptance \citep{shmueli_predictive_2011, kuhl_how_2021}.

The remaining paper is structured as follows: In Section \ref{sec:background}, we introduce some related work. We propose the new sparse model in Section \ref{sec:igann_sparse}. We test this model according to experiments described in Section \ref{sec:exp_design}, followed by a presentation and discussion of the results in Section \ref{sec:results}, and Section \ref{sec:discussion}, respectively.

\chapter{Conceptual Background and Related Work} \label{sec:background}

\section{Sparse Prediction Models}

When dealing with high dimensional data, it is often beneficial to use methods to decrease the number of features impacting the final prediction. Sparsity improves model interpretability because the human mind is not capable of processing the number of information units (i.e., features) that an ML model can process at a time. This limitation lies around $7\pm2$ information units \citep{sparsity_psych_2, rudin_stop_2019}. There are a number of techniques such as compression, where the model size is decreased, principal component analysis (PCA), which finds vectors orthogonal to each other, in order to mathematically represent the most information in the least features, and feature selection \citep{sparsity_FS_review}. While compression and PCA lead to model inputs which humans are unable to comprehend intuitively, feature selection methods simply reduce the number of features that a ML model bases its prediction on. This allows researchers and decision-makers to fully comprehend the model behavior \citep{sparsity_FS_review}.

One way of determining the relevance of a feature is to use some linear model to determine its predictive power with regards to the target variable. Lasso used as feature selector does this by fitting a linear regression with $L1$ regularization \citep{hastie2009elements}. Thereby, coefficients are pushed to zero in order to result in a sparse model where most of the input features can be discarded. These methods, however, fall short in cases where features naturally have a non-linear effect. For instance in the context of health analytics, neither a body temperature that is too low nor too high is healthy, thus, linear feature selectors can easily ignore the importance of such an oftentimes powerful feature.

\section{Interpretable Generalized Additive Neural Network (IGANN)}
Increases in complexity of neural networks or models such as gradient boosted decision trees have improved the predictive power in a trade off for interpretability, leading to so-called black-box models. To tackle this challenge, recently proposed methods have kept some of the core innovations from black-box models, but altered core parts to obtain fully interpretable models, such as GAMs \citep[e.g.,][]{GAM_EBM, IGANN_Kraus}. 

IGANN, a novel ML model belonging to the GAM family, fits shape functions using a boosted ensemble of neural networks, where each network represents an extreme learning machine (ELM), as illustrated in Figure~\ref{fig:igann_functionality}. ELMs are simple feed-forward neural networks that use a faster learning method than gradient-based algorithms \citep{huang2006extreme}. In detail, the training only includes updating the weights of the output layer and, thus, is equal to fitting a linear model. Overall, IGANN has been shown to produce smooth shape functions that can be easily comprehended by users. Furthermore, the way in which IGANN trains the networks makes it an interesting choice to introduce model sparsity in a non-linear fashion, which we describe in the following.

\chapter{IGANN Sparse} \label{sec:igann_sparse}

\begin{figure}[H]
\centering
  \includegraphics[width=0.4\linewidth]{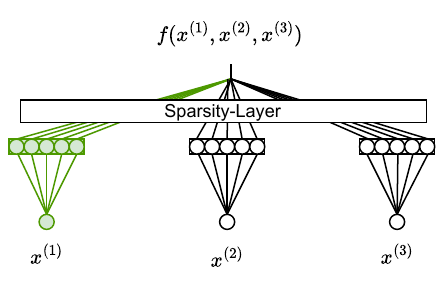}
  \caption{First ELM from IGANN Sparse with three features as input which includes the sparsity-layer. Each feature is processed by a sub-network of the whole ELM. For input $x^{(1)}$, the green part of the model highlights the corresponding sub-network.}
  \label{fig:igann_functionality}
\end{figure}

As described above, IGANN uses a sequence of ELMs to compute the GAM. In the following, we make use of this model choice to incorporate a sparsity-layer in the first ELM which allows to select the (potentially non-linear) most important features. Figure~\ref{fig:igann_functionality} illustrates this basic idea. 

For a fixed number of inputs $x^{(1)}, x^{(2)}, \ldots, x^{(m)}$, the ELM maps each input onto $k$ non-linear hidden activations, which we call $h^{(1)}_1, \ldots, h^{(1)}_k$, $\ldots$, $h^{(m)}_1, \ldots, h^{(m)}_k$. We denote the vectors that store the respective hidden activations by $h^{(i)}$, i.e., $h^{(i)} = [h^{(i)}_1, \ldots, h^{(i)}_k]$ and let $h$ represent all hidden activations, i.e., $h = [h^{(1)}, \ldots, h^{(m)}]$.

In the traditional IGANN model, the ELM is now trained by solving the linear problem

\begin{equation}
\label{equ:ELM}
    \min_\beta \mathcal{L}(y, \beta^T\,h)
\end{equation}
where $\beta$ denotes the coefficients in the last layer, $y$ the true target variable, and $\mathcal{L}$ describes the loss function, such as mean squared error or cross-entropy. As can be seen in Equation~\ref{equ:ELM}, the ELM thus merely solves a linear problem, yet the non-linear activations allow to capture highly non-linear effects \citep{IGANN_Kraus}. 
\vspace{-0.5cm}

This work makes use of this characteristic by introducing a sparsity-layer. Given the non-linear activations $h^{(1)}, \ldots, h^{(m)}$, where each $h^{(i)}$ represents a block of $k$ values, it's critical to ensure that the model remains interpretable and avoids overfitting by using only the most important blocks of activations. To achieve this, we introduce a sparsity-inducing step using the best-subset selection approach \citep{zhu2022abess}.

Considering each block $h^{(i)}$ as an individual subset, the best-subset selection aims to find the subset of blocks which, when used in the ELM model, results in the optimal balance between model fit and complexity. Mathematically, the problem can be extended from Equation~\ref{equ:ELM} as

\begin{equation}
\label{equ:ELM_Sparse}
    \min_{\beta, S} \mathcal{L}(y, \beta^T\,h_S),
\end{equation}

where $S$ denotes the selected subset of blocks from $h^{(1)}, \ldots, h^{(m)}$, and $h_S$ represents the hidden activations corresponding to this subset. The objective is to minimize the Bayesian Information Criterion ($BIC$) which is defined as

\begin{equation}
    BIC = |S|\, \ln(n) - 2 \ln(\hat{L}), 
\end{equation}
where $S$ is the set of all selected blocks, $|S|$ is the number of selected blocks (corresponding to features) by the model, $n$ is the number of observations, and $\hat{L}$ is the observed value of the negative loss function for the model. By using $BIC$ in the selection process, only the blocks that add significant explanatory power to the model's predictions are retained, resulting in a sparser and more interpretable model.

With this approach, the majority of blocks in $h_S$ can be set to zero, leaving only those blocks that contribute most significantly to the model's predictive power. This results in a sparse representation of the hidden activations.

\chapter{Experiment Design} \label{sec:exp_design}

\section{Datasets and Pre-Processing}
Our experiments are based on common, publicly available benchmark datasets presented in Table \ref{tab:datasets_overview}. The number of categorical features \textquote{cat} in these tables are measured after one-hot encoding. For preprocessing, we removed columns like IDs or for categorical features with more than 25 distinct values as is done in similar experiments \citep[e.g.,][]{zschech2022gam}. Furthermore, the standard scaler from scikit-learn is used for all numerical features.
The data is split in a 5-fold cross validation to evaluate the model's performance.

\begin{table}[ht]
\centering
\footnotesize
\setlength{\tabcolsep}{1pt}
\sisetup{group-digits=true, group-separator={,}, group-minimum-digits=4}
\begin{tabular}{p{5.4cm}S[table-format=6.0]rr@{\hspace{1em}}p{5.4cm}S[table-format=6.0]rr}
\toprule
\multicolumn{4}{c}{\textbf{Classification}} & \multicolumn{4}{c}{\textbf{Regression}} \\ 
\cmidrule(lr){1-4} \cmidrule(lr){5-8}
\textbf{Dataset} & \textbf{Samples} & \textbf{num} & \textbf{cat} & \textbf{Dataset} & \textbf{Samples} & \textbf{num} & \textbf{cat} \\
\midrule
college \citep{college_source} & 1000 & 4 & 10 & bike  \citep{fanaee-t_event_2014} & 17379 & 7 & 5 \\
churn \citep{IBM} & 7043 & 3 & 37 & wine \citep{Cortez2009} & 4898 & 11 & 0 \\
credit \citep{fico_source} & 10459 & 21 & 16 & productivity \citep{Imran_2019} & 1197 & 9 & 26 \\
income  \citep{Kohavi.1996} & 32561 & 6 & 59 & insurance  \citep{lantz_2015} & 1338 & 3 & 6 \\
bank  \citep{moro2014-driven_2014} & 45211 & 6 & 41 & crimes  \citep{redmond_data-driven_2002} & 1994 & 100 & 0 \\
airline  \citep{tj_klein_airline} & 103904 & 18 & 6 & farming \citep{gursewak_singh_sidhu_2021} & 3893 & 7 & 3 \\
recidivism  \citep{angin_machine_2016} & 7214 & 7 & 4 & house \citep{pace_sparse_1997} & 20640 & 8 & 0 \\
\bottomrule
\end{tabular}%
\caption{Overview of selected datasets covering classification ($y\in\{0,1\}$) and regression ($y\in\mathcal{R}$) tasks. Samples describes the number of observations recorded in the dataset (number of rows). Numerical (num) features and categorical (cat) features are the number of input columns representing numerical or categorical values, respectively}. Cat features are measured after one-hot encoding.
\label{tab:datasets_overview}
\end{table}

\section{Experiments}

Our first experiment compares the prediction quality of our sparse model versus an unconstrained IGANN model. This experiment assesses the trade-off between the quality of prediction and the level of sparsity.

Our second experiment tests the performance of the IGANN Sparse model for feature selection by comparing it to a lasso model for feature selection. The main metrics to evaluate on are the number of selected features, and the area under the receiver operating characteristic (AUROC) and root mean squared error (RMSE) for classification and regression, respectively.

Both experiments are repeated for 20 times with different random states in both data split and during model training, in order to gain a statistical distribution of the results, while maintaining reproducibility of the results. The statistical analysis is done using a Wilcoxon test \citep{Wilcoxon_Textbuch_Neuhaeuser}. The test assumes the $h0$ hypothesis of similar model performance. For the comparison in experiment one we consider similar model performance within a tolerance of one standard deviation, as a sparser model with comparable performance is better in fields where comprehensibility is required \citep{rudin_stop_2019}. The statistical analysis for the second experiment comparing IGANN Sparse and lasso as feature selectors will be conducted without tolerance.

\chapter{Results} \label{sec:results}

Table \ref{tab:performance_overview} shows the performance of IGANN Full and IGANN Sparse across 20 runs with a 5-fold cross validation. For both experiments, the results of the Wilcoxon tests are highlighted in the tables. In only three cases, the sparse model selected equal to or more than 75 \% of the datasets' total features, in one case as few as 4 \%.

The sparse model is considered superior if within one standard deviation of its non-sparse counterpart.
The comparison between the predictive performance of IGANN Full and Sparse shows that for 10 out of 14 datasets IGANN sparse is significantly better at p$\leq$0.01. Each of the statistical tests are based on 100 observations. 
Moreover, IGANN Sparse is significantly better in 11 out of 14 datasets at p$\leq$0.05.

\begin{figure}[h!]
  \centering
  \begin{subfigure}[b]{0.47\linewidth}
    \includegraphics[width=\linewidth]{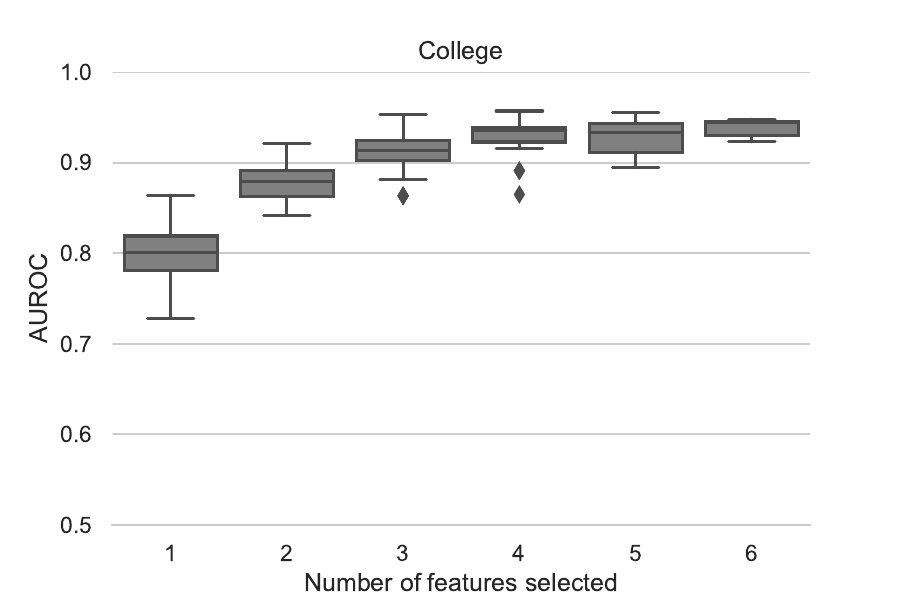}
  \end{subfigure}
  \hfill
  \begin{subfigure}[b]{0.47\linewidth}
    \includegraphics[width=\linewidth]{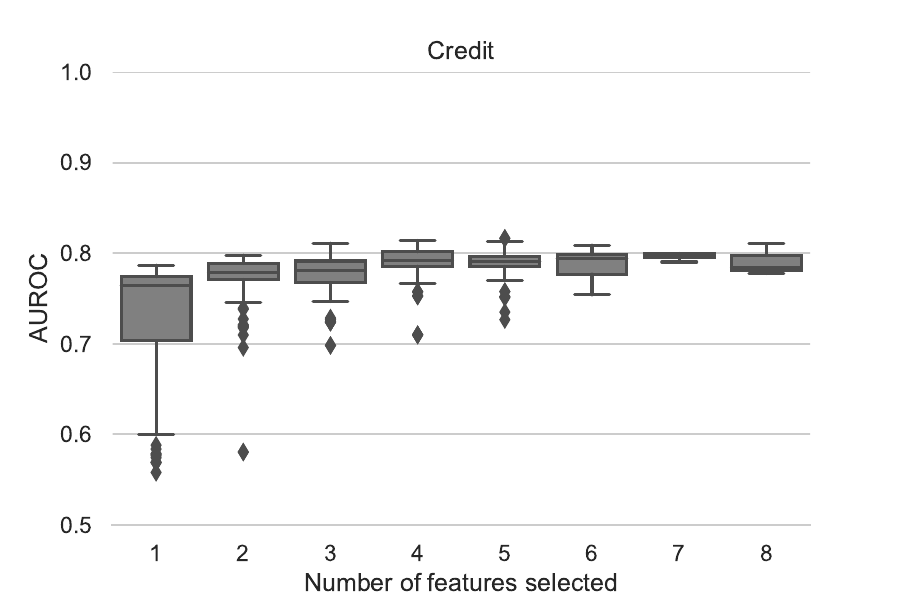}
  \end{subfigure}
  \caption{Impact of the number of features selected on model performance for the college and credit} dataset as measured using the AUROC score (closer to 1 is better).
  \label{fig:numb_feats_auroc}
\end{figure}

Figure \ref{fig:numb_feats_auroc} exemplary shows the performance for two datasets (college and credit) for varying numbers of selected features. Already with as few as four features we find very promising predictive performance which is not further improved using more features.

As a feature selector, IGANN Sparse performed better than lasso in 9 out of 14 cases. In three of our seven classification datasets and in two of our seven regression datasets, lasso performed better than IGANN Sparse. Making IGANN Sparse the better feature selector in the majority of datasets for both classification and regression tasks.

\begin{table*}[htbp]
\footnotesize
\centering
\setlength{\tabcolsep}{1pt}

\begin{tabular}{@{}lclclcllllclcc@{}}
\toprule
& \multicolumn{6}{c}{Classification} & \multicolumn{6}{c}{Regression} \\
\cmidrule(lr){2-7} \cmidrule(l){8-13}
Dataset & \multicolumn{2}{c}{IGANN Full} & \multicolumn{3}{c}{IGANN Sparse} & & Dataset & \multicolumn{2}{c}{IGANN Full} & \multicolumn{3}{c}{IGANN Sparse} \\
\cmidrule(lr){2-3} \cmidrule(lr){4-6} \cmidrule(lr){9-10} \cmidrule(l){11-13}
& AUROC & $\pm$ SD & AUROC & $\pm$ SD & \# Features & & & RMSE & $\pm$ SD & RMSE & $\pm$ SD & \# Features \\
\midrule
college & 0.863 & $\pm$ 0.022 & 0.852 & $\pm$ 0.025$^{**}$ & 72.0 \% & & bike & 0.766 & $\pm$ 0.006 & 0.768 & $\pm$ 0.007$^{**}$ & 80.4 \% \\
churn & 0.722 & $\pm$ 0.012 & 0.711 & $\pm$ 0.013$^{**}$ & 51.5 \% & & wine & 0.901 & $\pm$ 0.015 & 0.914 & $\pm$ 0.016$^{*}$ & 34.8 \% \\
credit & 0.731 & $\pm$ 0.009 & 0.725 & $\pm$ 0.016$^{**}$ & 44.8 \% & & productivity & 0.896 & $\pm$ 0.032$^{**}$ & 0.960 & $\pm$ 0.038 & 62.3 \% \\
income & 0.775 & $\pm$ 0.006$^{**}$ & 0.706 & $\pm$ 0.024 & 51.1 \% & & insurance & 0.706 & $\pm$ 0.020 & 0.707 & $\pm$ 0.020$^{**}$ & 68.0 \% \\
bank & 0.587 & $\pm$ 0.005 & 0.584 & $\pm$ 0.006$^{**}$ & 90.1 \% & & crimes & 0.771 & $\pm$ 0.026 & 0.778 & $\pm$ 0.026$^{**}$ & 4.0 \% \\
airline & 0.933 & $\pm$ 0.002$^{**}$ & 0.929 & $\pm$ 0.002 & 64.6 \% & & farming & 0.815 & $\pm$ 0.020 & 0.822 & $\pm$ 0.020$^{**}$ & 56.0 \% \\
recidivism & 0.685 & $\pm$ 0.010 & 0.680 & $\pm$ 0.014$^{**}$ & 51.4 \% & & house & 0.733 & $\pm$ 0.006 & 0.735 & $\pm$ 0.005$^{**}$ &  78.6 \%\\ 
\bottomrule
\end{tabular}
\caption{Performance comparison of classification and regression datasets, showing AUROC and RMSE results as well as standard deviations (SD) of the standard IGANN model compared to the sparse model with the average percentage of selected features out of the input features described in Table \ref{tab:datasets_overview}, including both categorical as well as numerical features. For classification, values closer to 1 are better, and for regression, lower values are better. Significant differences using the Wilcoxon signed rank test are marked with $** p \leq 0.01$, $* p \leq 0.05$ at the respectively better model. We consider the sparser model to be better if it performs within at least one standard deviation of the full model, due to its easier comprehensibility.}
\label{tab:performance_overview}
\end{table*}

\chapter{Discussion \& Future Work} \label{sec:discussion}
This research introduced IGANN Sparse, a novel model designed for sparsity and interpretability in predictive analytics projects. We validated its usage as both, a predictive model and a feature selector on a variety of datasets. Despite resulting in much sparser models with as little as 4 \% of the input features, we showed that IGANN Sparse obtains competitive results in more than 75 \% of our tested datasets. Additionally, IGANN Sparse outperformed traditional feature selectors in the majority of cases. 
In future work, we plan to include further state-of-the-art (black-box) models for a broader comparison, as well as more diverse datasets with different properties. In this way, we aim to better understand our model's merits and limitations and explore its generalizability to different types of data, tasks, and contexts. Further, we want to analyse the effect of including pairwise interactions to potentially improve predictive performance.

A key aspect of IGANN Sparse is its role as an exploratory tool for IS researchers, particularly capable of revealing non-linear relationships within data, an area not fully explored by existing methods. This capability positions IGANN Sparse not only as a predictive or feature-reducing model, but also as a means for deeper data understanding. Future research can leverage the exploratory strengths of IGANN Sparse in conjunction with domain-specific knowledge to validate theoretical models or to hypothesize new relationships. This could be especially transformative in interdisciplinary IS research, where the fusion of methodological robustness with domain expertise is essential for innovation.

Our assessment of interpretability primarily considered feature reduction from a mathematical perspective. While this is a critical element \citep{wang_sparsity_required}, it alone does not capture the full spectrum of model transparency. Interpretability extends beyond numerical simplicity to the model's ability to convey understanding and actionable insights to researchers and decision makers. 
To this end, we plan to conduct user studies focused on evaluating the practical interpretability of IGANN Sparse, aiming to capture comprehensive, qualitative feedback from domain experts on its effectiveness and applicability in real-world scenarios. These studies are designed to explore how well different stakeholders can grasp the underlying model mechanics, apply its predictions to complex settings, and trust its recommendations in their professional environment.

In conclusion, our research shows that simplification in prediction models does not require compromising performance. IGANN Sparse shows the potential of machine learning models that can achieve the critical balance between simplicity and accuracy that is decisive to practical understanding and application. We believe that the principles and results of this study will make a valuable contribution to the IS discourse and inspire further research in interpretable modeling.

\printbibliography

@article{huang2006extreme,
  title={Extreme learning machine: {Theory} and applications},
  author={Huang, Guang-Bin and Zhu, Qin-Yu and Siew, Chee-Kheong},
  journal={Neurocomputing},
  volume={70},
  number={1-3},
  pages={489--501},
  year={2006},
  publisher={Elsevier}
}

@inproceedings{zschech2022gam,
	title = {{GAM}(e) Change or Not? {An} Evaluation of Interpretable Machine Learning Models Based on Additive Model Constraints},
	booktitle = {Proceedings of the 30th {European} {Conference} on {Information} {Systems} ({ECIS})},
	author = {Zschech, Patrick and Weinzierl, Sven and Hambauer, Nico and Zilker, Sandra and Kraus, Mathias},
	year = {2022},
}

@article{GAM_Hastie_Tibishirani,
author = {Trevor Hastie and Robert Tibshirani},
title = {{Generalized Additive Models}},
volume = {1},
journal = {Statistical Science},
number = {3},
publisher = {Institute of Mathematical Statistics},
pages = {297 -- 310},
year = {1986},
}

@article{GAM_EBM,
  author       = {Harsha Nori and
                  Samuel Jenkins and
                  Paul Koch and
                  Rich Caruana},
  title        = {InterpretML: {A} Unified Framework for Machine Learning Interpretability},
  year         = {2019},
  eprinttype    = {arXiv},
  eprint       = {1909.09223},
  bibsource    = {dblp computer science bibliography, https://dblp.org}
}

@article{GAM_GAMI-Net,
title = {GAMI-Net: An explainable neural network based on generalized additive models with structured interactions},
journal = {Pattern Recognition},
volume = {120},
pages = {108192},
year = {2021},
issn = {0031-3203},
author = {Zebin Yang and Aijun Zhang and Agus Sudjianto}
}

@article{IGANN_Kraus,
title = {Interpretable generalized additive neural networks},
journal = {European Journal of Operational Research},
year = {2023},
issn = {0377-2217},
author = {Mathias Kraus and Daniel Tschernutter and Sven Weinzierl and Patrick Zschech},
}

@article{lundberg2017unified,
  title={A unified approach to interpreting model predictions},
  author={Lundberg, Scott M and Lee, Su-In},
  journal={Advances in neural information processing systems},
  volume={30},
  year={2017}
}

@book{hastie2009elements,
  title={The elements of statistical learning: data mining, inference, and prediction},
  author={Hastie, Trevor and Tibshirani, Robert and Friedman, Jerome H and Friedman, Jerome H},
  volume={2},
  year={2009},
  publisher={Springer}
}

@inproceedings{ribeiro2016should,
  title={"Why should i trust you?" Explaining the predictions of any classifier},
  author={Ribeiro, Marco Tulio and Singh, Sameer and Guestrin, Carlos},
  booktitle={Proceedings of the 22nd ACM SIGKDD international conference on knowledge discovery and data mining},
  pages={1135--1144},
  year={2016}
}

@misc{zhu2022abess,
      title={abess: A Fast Best Subset Selection Library in Python and R}, 
      author={Jin Zhu and Xueqin Wang and Liyuan Hu and Junhao Huang and Kangkang Jiang and Yanhang Zhang and Shiyun Lin and Junxian Zhu},
      year={2022},
      eprint={2110.09697},
      archivePrefix={arXiv}
}

@article{wang_sparsity_required,
	title = {In Pursuit of Interpretable, Fair and Accurate Machine Learning for Criminal Recidivism Prediction},
	volume = {39},
	issn = {1573-7799},
	pages = {519--581},
	number = {2},
	journaltitle = {Journal of Quantitative Criminology},
	shortjournal = {Journal of Quantitative Criminology},
	author = {Wang, Caroline and Han, Bin and Patel, Bhrij and Rudin, Cynthia},
	date = {2023},
}

@article{rudin_stop_2019,
	title = {Stop explaining black box machine learning models for high stakes decisions and use interpretable models instead},
	volume = {1},
	issn = {2522-5839},
	pages = {206--215},
	number = {5},
	journaltitle = {Nature Machine Intelligence},
	shortjournal = {Nature Machine Intelligence},
	author = {Rudin, Cynthia},
	date = {2019},
}

@article{esteva_guide_2019,
	title = {A guide to deep learning in healthcare},
	volume = {25},
	issn = {1546-170X},
	pages = {24--29},
	number = {1},
	journaltitle = {Nature Medicine},
	shortjournal = {Nature Medicine},
	author = {Esteva, Andre and Robicquet, Alexandre and Ramsundar, Bharath and Kuleshov, Volodymyr and {DePristo}, Mark and Chou, Katherine and Cui, Claire and Corrado, Greg and Thrun, Sebastian and Dean, Jeff},
	date = {2019},
}

@article{sparsity_psych_2,
  title={The magical number seven, plus or minus two: Some limits on our capacity for processing information.},
  author={Miller, George A},
  journal={Psychological review},
  volume={63},
  number={2},
  pages={81},
  year={1956},
  publisher={American Psychological Association}
}

@ARTICLE{sparsity_FS_review,
  author={Gui, Jie and Sun, Zhenan and Ji, Shuiwang and Tao, Dacheng and Tan, Tieniu},
  journal={IEEE Transactions on Neural Networks and Learning Systems}, 
  title={Feature Selection Based on Structured Sparsity: A Comprehensive Study}, 
  year={2017},
  volume={28},
  number={7},
  pages={1490-1507},
}

@Inbook{Wilcoxon_Textbuch_Neuhaeuser,
author="Neuh{\"a}user, Markus",
editor="Lovric, Miodrag",
title="Wilcoxon--Mann--Whitney Test",
bookTitle="International Encyclopedia of Statistical Science",
year="2011",
publisher="Springer Berlin Heidelberg",
address="Berlin, Heidelberg",
pages="1656--1658",
}

@misc{college_source,
	title={Go To College Dataset},
	url={https://www.kaggle.com/datasets/saddamazyazy/go-to-college-dataset},
	publisher={Kaggle},
	author={Saddam Sinatrya Jalu Mukti},
	year={2022}
}

@misc{fico_source,
	title={Explainable Machine Learning Challenge},
	author={{Fair Isaac Corporation}},
	year={2018}
}

@misc{tj_klein_airline,
	title={Airline Passenger Satisfaction},
	url={https://www.kaggle.com/datasets/teejmahal20/airline-passenger-satisfaction},
	publisher={Kaggle},
	author={TJ Klein},
	year={2020}
}

@misc{gursewak_singh_sidhu_2021,
	title={Crab Age Prediction},
	url={https://www.kaggle.com/dsv/2834512},
	DOI={10.34740/KAGGLE/DSV/2834512},
	publisher={Kaggle},
	author={Gursewak Singh Sidhu},
	year={2021}
}

@article{redmond_data-driven_2002,
	title = {A data-driven software tool for enabling cooperative information sharing among police departments},
	volume = {141},
	number = {3},
	journal = {European Journal of Operational Research},
	author = {Redmond, Michael and Baveja, Alok},
	year = {2002},
	pages = {660--678},
}

@article{fanaee-t_event_2014,
	title = {Event labeling combining ensemble detectors and background knowledge},
	volume = {2},
	number = {2},
	journal = {Progress in Artificial Intelligence},
	author = {Fanaee-T, Hadi and Gama, Joao},
	year = {2014},
	pages = {113--127}
}

@article{pace_sparse_1997,
	title = {Sparse spatial autoregressions},
	volume = {33},
	number = {3},
	journal = {Statistics \& Probability Letters},
	author = {Pace, Kelley R. and Barry, Ronald},
	year = {1997},
	pages = {291--297},
}

@inproceedings{Imran_2019,
year = 2019,
publisher = {{IEEE}},
author = {Abdullah Al Imran and Md Nur Amin and Md Rifatul Islam Rifat and Shamprikta Mehreen},
title = {Deep Neural Network Approach for Predicting the Productivity of Garment Employees},
booktitle = {2019 6th International Conference on Control, Decision and Information Technologies ({CoDIT})}
}

@book{lantz_2015, place={Birmingham}, title={Machine learning with R: Learn how to use R to apply powerful machine learning methods and gain an insight into real-world applications}, publisher={Packt Publ.}, author={Lantz, Brett}, year={2015}}

@article{moro2014-driven_2014,
	title = {A data-driven approach to predict the success of bank telemarketing},
	volume = {62},
	journal = {Decision Support Systems},
	author = {Moro, Sérgio and Cortez, Paulo and Rita, Paulo},
	year = {2014},
	pages = {22--31},
}

@misc{IBM,
  author = {IBM},
  title = {{Telco customer churn}},
  howpublished = "\url{https://community.ibm.com/community/user/businessanalytics/blogs/steven-macko/2019/07/11/telco-customer-churn-1113}",
  year = {2019}
}

@inproceedings{Kohavi.1996,
author = {Kohavi, Ron},
title = {Scaling up the Accuracy of Naive-Bayes Classifiers: A Decision-Tree Hybrid},
year = {1996},
publisher = {AAAI Press},
booktitle = {Proceedings of the Second International Conference on Knowledge Discovery and Data Mining},
pages = {202–207},
numpages = {6},
location = {Portland, Oregon},
series = {KDD'96}
}

@article{Cortez2009,
title = {Modeling wine preferences by data mining from physicochemical properties},
journal = {Decision Support Systems},
volume = {47},
number = {4},
pages = {547-553},
year = {2009},
note = {Smart Business Networks: Concepts and Empirical Evidence},
issn = {0167-9236},
author = {Paulo Cortez and António Cerdeira and Fernando Almeida and Telmo Matos and José Reis}
}

@incollection{angin_machine_2016,
  title={Machine bias},
  author={Angwin, Julia and Larson, Jeff and Mattu, Surya and Kirchner, Lauren},
  booktitle={Ethics of Data and Analytics},
  pages={254--264},
  year={2016},
  publisher={Auerbach Publications}
}

@article{shmueli_predictive_2011,
	title = {Predictive {Analytics} in {Information} {Systems} {Research}},
	volume = {35},
	issn = {02767783},
	doi = {10.2307/23042796},
	number = {3},
	journal = {MIS Quarterly},
	author = {{Shmueli} and {Koppius}},
	year = {2011},
	pages = {553},
}

@article{kuhl_how_2021,
	title = {How to {Conduct} {Rigorous} {Supervised} {Machine} {Learning} in {Information} {Systems} {Research}: {The} {Supervised} {Machine} {Learning} {Report} {Card}},
	volume = {48},
	issn = {15293181},
	number = {1},
	journal = {Communications of the Association for Information Systems},
	author = {Kühl, Niklas and Hirt, Robin and Baier, Lucas and Schmitz, Björn and Satzger, Gerhard},
	year = {2021},
	pages = {589--615},

}

@article{bauer_explained_2023,
	title = {Expl({AI})ned: {The} {Impact} of {Explainable} {Artificial} {Intelligence} on {Users}’ {Information} {Processing}},
	issn = {1047-7047},
	journal = {Information Systems Research},
	author = {Bauer, Kevin and von Zahn, Moritz and Hinz, Oliver},
	year = {2023},
}

@article{kim_rolex_2023,
	title = {{ROLEX}: {A} {Novel} {Method} for {Interpretable} {Machine} {Learning} {Using} {Robust} {Local} {Explanations}},
	volume = {47},
	issn = {ISSN 0276-7783/ISSN 2162-9730},
	shorttitle = {{ROLEX}},
	number = {3},
	journal = {Management Information Systems Quarterly},
	author = {Kim, Buomsoo R. and Srinivasan, Karthik and Kong, Sung Hye and Kim, Jung Hee and Shin, Chan Soo and Ram, Sudha},
	year = {2023},
	pages = {1303--1332},
}

@techreport{broniatowski,
	address = {Gaithersburg, MD},
	title = {Psychological foundations of explainability and interpretability in artificial intelligence},
	number = {NIST IR 8367},
	institution = {National Institute of Standards and Technology (U.S.)},
	author = {Broniatowski, David A},
	year = {2021},
	pages = {NIST IR 8367},
}

\end{document}